\documentclass{article}
\usepackage{graphicx} 
\usepackage{authblk}

\title{A Nasal Cytology Dataset for Object Detection and Deep Learning}

\author[1]{ Mauro Camporeale }
\author[1]{ Giovanni Dimauro }
\author[3]{ Matteo Gelardi }
\author[2]{ Giorgia Iacobellis }
\author[2]{ Mattia Sebastiano Ladisa }
\author[4]{ Sergio Latrofa }
\author[1]{ Nunzia Lomonte}

\affil[1]{Department of Computer Science, University of Bari}
\affil[2]{Department of Computer Science, University of Turin}
\affil[3]{Department of Otolaryngology, University Hospital of Foggia}
\affil[4]{Department of Computer Science, University of Pisa}

\date{March 2024}

\usepackage[sorting=none]{biblatex} 

\usepackage{caption}
\usepackage{subcaption}
\usepackage{verbatim}
\usepackage{multirow}

\begin{document}

\maketitle

\begin{abstract}
Nasal Cytology is a new and efficient clinical technique to diagnose rhinitis and allergies that is not much widespread due to the time-consuming nature of cell counting; that is why AI-aided counting could be a turning point for the diffusion of this technique.
In this article we present the first dataset of rhino-cytological field images: the NCD (Nasal Cytology Dataset), aimed to train and deploy Object Detection models to support physicians and biologists during clinical practice. The real distribution of the cytotypes, populating the nasal mucosa has been replicated, sampling images from slides of clinical patients, and manually annotating each cell found on them. The correspondent object detection task presents nontrivial issues associated with the strong class imbalancement, involving the rarest cell types. This work contributes to some of open challenges by presenting a novel machine learning-based approach to aid the automated detection and classification of nasal mucosa cells: the DETR \cite{DETR} and YOLO \cite{YOLOV8} models shown good performance in detecting cells and classifying them correctly, revealing great potential to accelerate the work of rhinology experts


\end{abstract}

\section{Introduction}

In recent years, Artificial Intelligence (AI) algorithms pervasively spread among many aspects of everyday life: health and medicine were no exceptions \cite{AI-PEDIATRIC, DL-ELECTROMIOGRAPHIC, ARHYTMIA_CLASSIFICATION}. In particular Deep Learning (DL) models were applied with great success to several clinical tasks concerning diagnosis by images \cite{TRANSFER-LEARNING-CERVIX-CYTO, CERVICAL-CANCER-STAINING}. One of these clinical tasks is to use AI to tackle cytology, generally defined as the practice of observing and analyzing cells within specimens sampled from some organ or tissue, to detect anomalies and possible ongoing diseases \cite{CYTO-ATLAS}. 
Thanks to deep architectures, several cytological pipelines were successfully learned and automatized by AI algorithms \cite{EMERGING-CYTO, COMP-CYTOLOGY, DEEPCILIA, DL-CYTOLOGY-SURVEY,AUTO-CYTO-REVIEW}, providing important support instruments to doctors on the ward. Several examples involving different organs/tissues and pathologies can be provided: cervix liquid cytology to diagnose and prevent brain cancer \cite{LIQUID-YOLO, NUCLEI-CERVICAL-CYTOLOGY, PRE-DIAGNOSIS-ML},  bronchoalveolar lavage fluid cytology \cite{AUTOMATED-BRONCOALVEOLAR} to detect various respiratory illness, blood cytology \cite{CYTO-BLOOD} to detect hematologic disorders and urine cytology \cite{URINE-CYTOLOGY} for the identification of atypical and malignant cells providing risk profiles for oncological diseases.
Similar works in the world of veterinary science can be found as well, for example, \cite{PRE-DIAGNOSIS-ML} and \cite{COW-CITOLOGY} concerning automated cytological practices respectively for dog skin pathologies and endometriosis in dairy cows. 

Although good results obtained with those kinds of approaches, an important bottleneck is still represented by the lack of openly available data related to experimental papers, mainly due to law restrictions, patient privacy issues or clinical/academic policies \cite{PRIVACY-ML-MED, DL-CYTO-CHALLENGES}; in particular, for cytology some public datasets exist, mainly concerning cervical tissue \cite{REVIEW-DL-CYTO-CERVIX}. An important update has recently been given by the Live-Cell dataset \cite{LIVE-CELL} and by the EVICAN dataset \cite{EVICAN}, both consisting of (unlabeled) microscope images where cells are annotated within a rectangle called a "bounding box", representing a general purpose benchmark for cell segmentation algorithms, like \cite{CELLPOSE}.
Those data have a great potential, being exploitable to pre-train DL models, i.e. learning a transferable embedding space \cite{REVIEW-DL-CYTO-CERVIX} to subsequently fine-tune on some new tissue-specific tasks for whose only a limited number of samples is available.

This work contributes to some of open challenges by presenting a novel machine learning-based approach to aid the automated detection and classification of nasal mucosa cells: the DETR [1] and YOLO [2] models shown good performance in detecting cells and classifying them correctly, revealing great potential to accelerate the work of rhinology experts.
A valuable contribution of this article is the publication of the nasal citology datased (NCD), a novel dataset, composed of more than 10.000 instances of cell of the nasal mucosa, distributed throughout 500 images. We provide a this new open dataset as a benchmark to train and deploy Computer Vision or AI models focused on the cytological practice concerning nasal mucosa. This dataset allows to explore both the cell detection task and cell classification task, which will be from now on referred to as the cell recognition task, hopefully contributing to the development of AI-aided medicine research \cite{MEDICAL-PYISICS-IN-ITALY} and to the still large unexplored potentialities of AI applied to Rhinology \cite{ML-RYNOLOGY, NASAL-DL, MEDICAL-DATA-SCIENCE-RYNOLOGY}.

In general, it is of little significance to compare the results obtained with different datasets and experimental sets that are not homogeneous. Instead of offering comparisons with other studies, it is useful to define and propose a common experimental basis. We think that this study can be considered as a new starting point for the community working on this topic and can promote further studies with the aim of obtaining the best possible results and package a very useful system for doctors. 

This document is organized as follows: in the next subsection, a background about nasal cytology and cells of nasal mucosa is provided. In section 2, we describe the dataset construction pipeline, while in section 3 are briefly described the chosen models and metrics. In Section \ref{sec:experiments and results} the experiments are described: they were carried on with the DETR and YOLO models to be used as an initial benchmark on both the Cell Detection and the Cell Recognition Tasks. Finally, section \ref{sec:discussion} and \ref{sec:conclusion} respectively discuss the results and draw the conclusion of this work.

\subsection{Nasal Cytology}

Nasal Cytology, or Rhinology, is the subfield of otolaryngology, focused on the microscope observation of samples of the nasal mucosa, aimed to recognize cells of different types, to spot and diagnose ongoing pathologies \cite{CYTO-ATLAS, ALLERGIC-RHINITIS}.

Such methodology can claim good accuracy in diagnosing rhinitis and infections, being very cheap and accessible without any instrument more complex than a microscope, even optical ones. Mucosa samples are taken non-invasively, just using a simple swab, to be then smeared onto a glass (fixation) and coloured with staining, in our case the May-Grunwald-Giemsa, before being observed at the microscope. 

\begin{figure}
     \centering
     \begin{subfigure}[b]{0.18\textwidth}
         \centering
         \includegraphics[width=\textwidth]{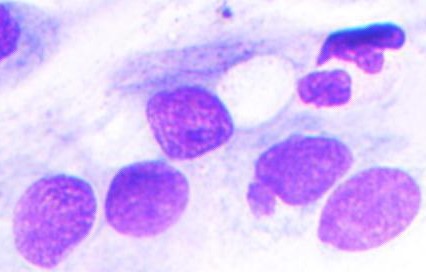}
         \caption{Epithelial}
         \label{1a}
     \end{subfigure}
     \hfill
     \begin{subfigure}[b]{0.18\textwidth}
         \centering
         \includegraphics[width=\textwidth]{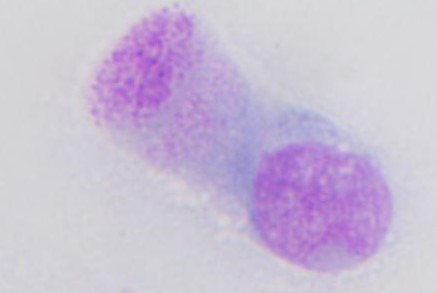}
         \caption{Ciliated}
         \label{1b}
     \end{subfigure}
     \hfill
     \begin{subfigure}[b]{0.18\textwidth}
         \centering
         \includegraphics[width=\textwidth]{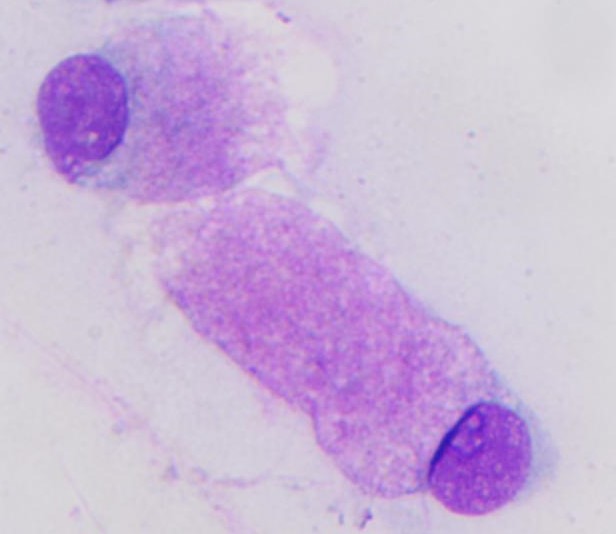}
         \caption{Muciparous}
         \label{1c}
     \end{subfigure}
      \hfill
     \begin{subfigure}[b]{0.18\textwidth}
         \centering
         \includegraphics[width=\textwidth]{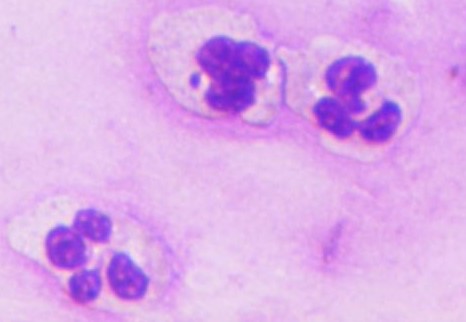}
         \caption{Neutrophil}
         \label{1d}
     \end{subfigure}
      \hfill
     \begin{subfigure}[b]{0.18\textwidth}
         \centering
         \includegraphics[width=\textwidth]{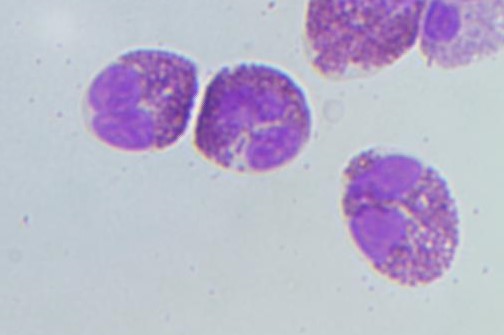}
         \caption{Eosinophil}
         \label{1e}
     \end{subfigure}
          \begin{subfigure}[b]{0.2\textwidth}
         \centering
         \includegraphics[width=\textwidth]{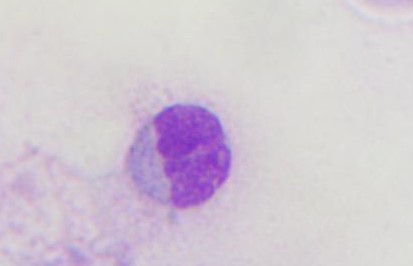}
         \caption{Lymphocyte}
         \label{1f}
     \end{subfigure}
     \hfill
     \begin{subfigure}[b]{0.16\textwidth}
         \centering
         \includegraphics[width=\textwidth]{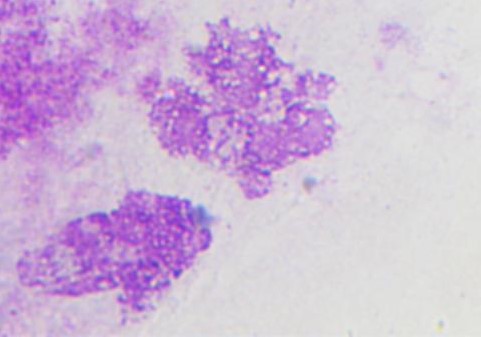}
         \caption{Mast Cell}
         \label{1g}
     \end{subfigure}
     \hfill
     \begin{subfigure}[b]{0.18\textwidth}
         \centering
         \includegraphics[width=\textwidth]{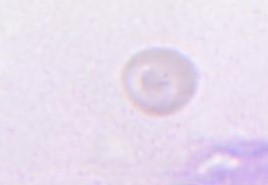}
         \caption{Ematia}
         \label{1h}
     \end{subfigure}
      \hfill
     \begin{subfigure}[b]{0.18\textwidth}
         \centering
         \includegraphics[width=\textwidth]{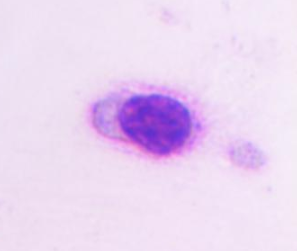}
         \caption{Metaplastic}
         \label{1i}
     \end{subfigure}
      \hfill
     \begin{subfigure}[b]{0.18\textwidth}
         \centering
         \includegraphics[width=\textwidth]{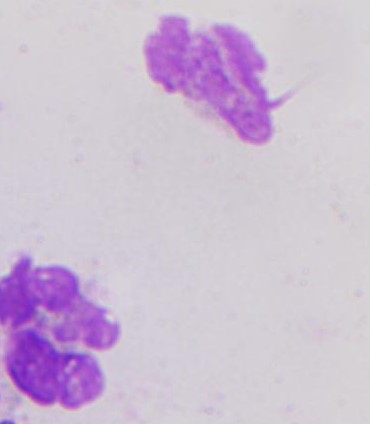}
         \caption{Artifact}
         \label{1j}
     \end{subfigure}
        \caption{Examples of observable items with MGG staining.}
        \label{fig:cytotipes}
\end{figure}

Cells of the nasal mucosa are distinguishable by some specific characteristic proper of each cytotype. In the current study, the considered cells are: 
\begin{description}

\item{\textbf{Epithelial Cells}}: main components of nasal mucosa, constituting ~80\% of the observed cytotype (Fig. \ref{1a}) in health patients' exams.  Their presence is not associated with ongoing pathologies. 

\item {\textbf{Ciliated cells}}: into the epithelium cells family we can find also those sub-type, characterized by their "tailed-like" shape (Fig. \ref{1b}).

\item \textbf{Metaplastic cells}: another sub-type of epithelial cells, characterized by their round shape. Their presence is usually associated with ongoing inflammatory reaction. 

\item{\textbf{Muciparous}}: calciform mucous-secreting cells (Fig. \ref{1c}) characterized by a bilobed shape with chromatin reinforced membrane.  The increase of muciparous cells results in increased mucus production, a symptom of nasal pathologies with chronic trends, like, in example, dust mites allergies.

\item{\textbf{Neutrophils}}: granulocytes with several nucleoli and a round shape (Fig. \ref{1d}). Their main function is the phagocytosis of germs. An increase in their number should always be kept under control as an immune response indicator.        
    
\item{\textbf{Eosinophils}}: polynuclear granulocytes, slightly large than neutrophils (Fig. \ref{1e}). The MGG staining tends to highlight eosinophil grains with an orange color. Allergic diseases are associated with an increase in their population (15-30\%). 

\item{\textbf{Lymphocytes}}: white blood cells (Fig. \ref{1f}) responsible for the immune response. Their large nucleus is surrounded by a thin cytoplasmatic "light blue" rim. 

\item{\textbf{Mast-cells}}: large oval cells having their nuclei covered with basophil granules of intense color (Fig. \ref{1g}). Their presence in the nasal mucosa  is caused by ongoing allergies. 

\item{\textbf{Ematia (\textit{Erythrocyte})}}: red blood cells whose occurrence in rhinological specimen may be due to pathologies or previous internal nose wounds, or even to small blood losses during the smear process (Fig.\ref{1h}).

\item{\textbf{Artifacts}}: with this name, we intended an object with morphology similar to the one of a cell but not being one, that can be considered an artifact. Examples of artifacts may be pollen pieces (Fig. \ref{1j})  or dirt spots on the slide. 

\end{description}

\section{Dataset Construction}
\label{sec:dataset construction}

The construction of the NCD dataset is the result of intense work and collaboration between otolaryngologists and computer scientists who, convinced of the great contribution that artificial intelligence can make to this branch of medicine, decided to make material available to the scientific community to allow them to challenge and confront each other in this new application field.

\begin{figure}
     \centering
     \begin{subfigure}[b]{0.45\textwidth}
         \centering
         \includegraphics[width=\textwidth]{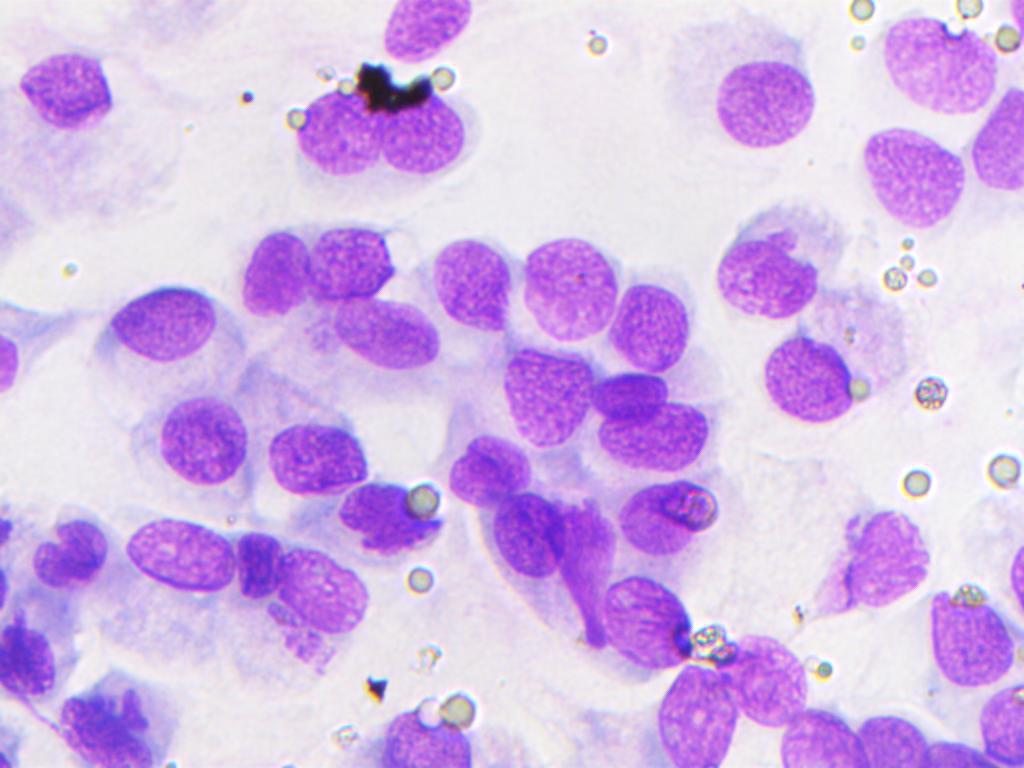}
         \caption{A field with pollen pieces}
         \label{2a}
     \end{subfigure}
     \hfill
     \begin{subfigure}[b]{0.45\textwidth}
         \centering
         \includegraphics[width=\textwidth]{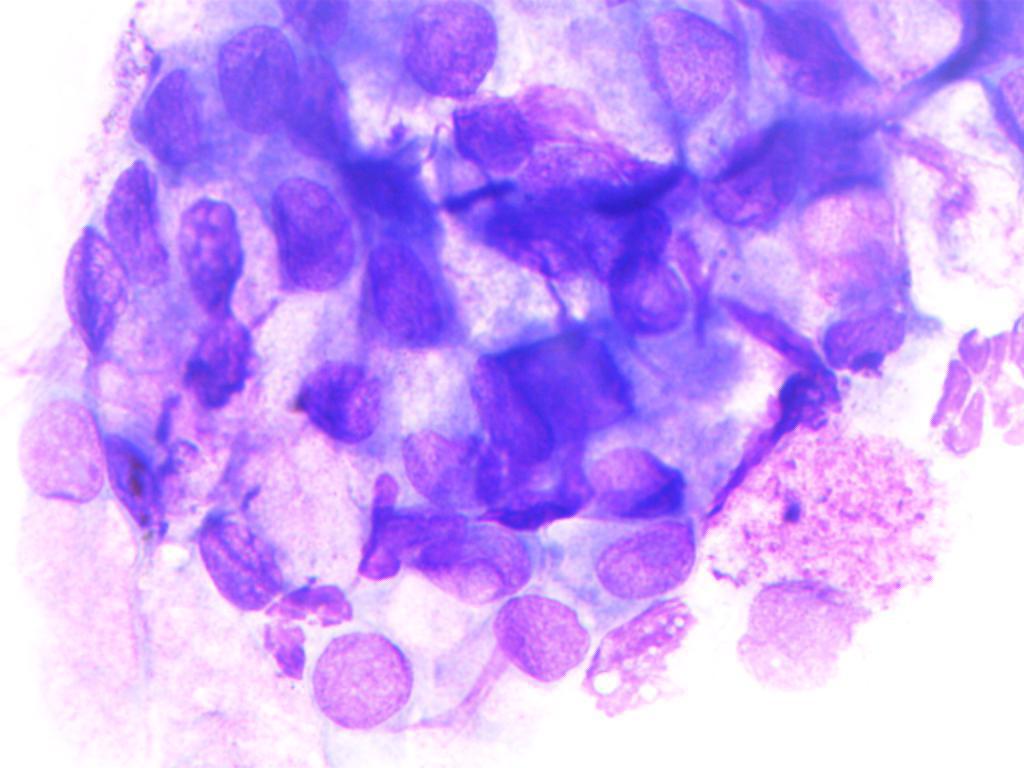}
         \caption{Dense cluster of cells}
         \label{2b}
     \end{subfigure}
          \begin{subfigure}[b]{0.45\textwidth}
         \centering
         \includegraphics[width=\textwidth]{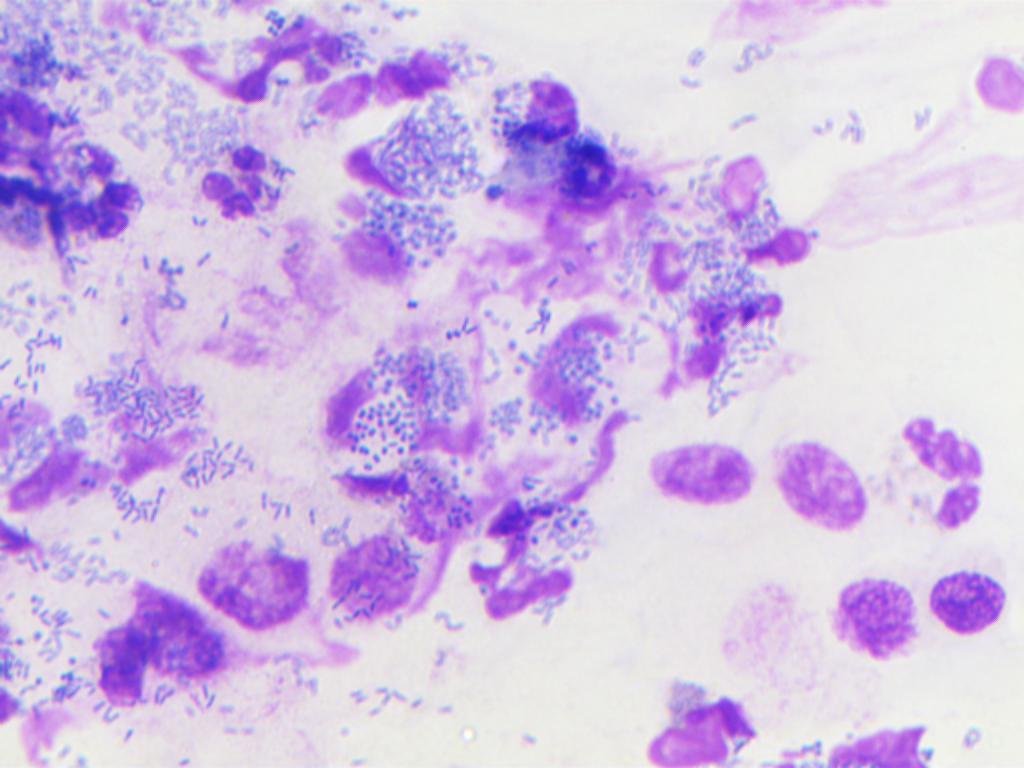}
         \caption{A bacteria colony}
         \label{2c}
     \end{subfigure}
     \hfill
     \begin{subfigure}[b]{0.45\textwidth}
         \centering
         \includegraphics[width=\textwidth]{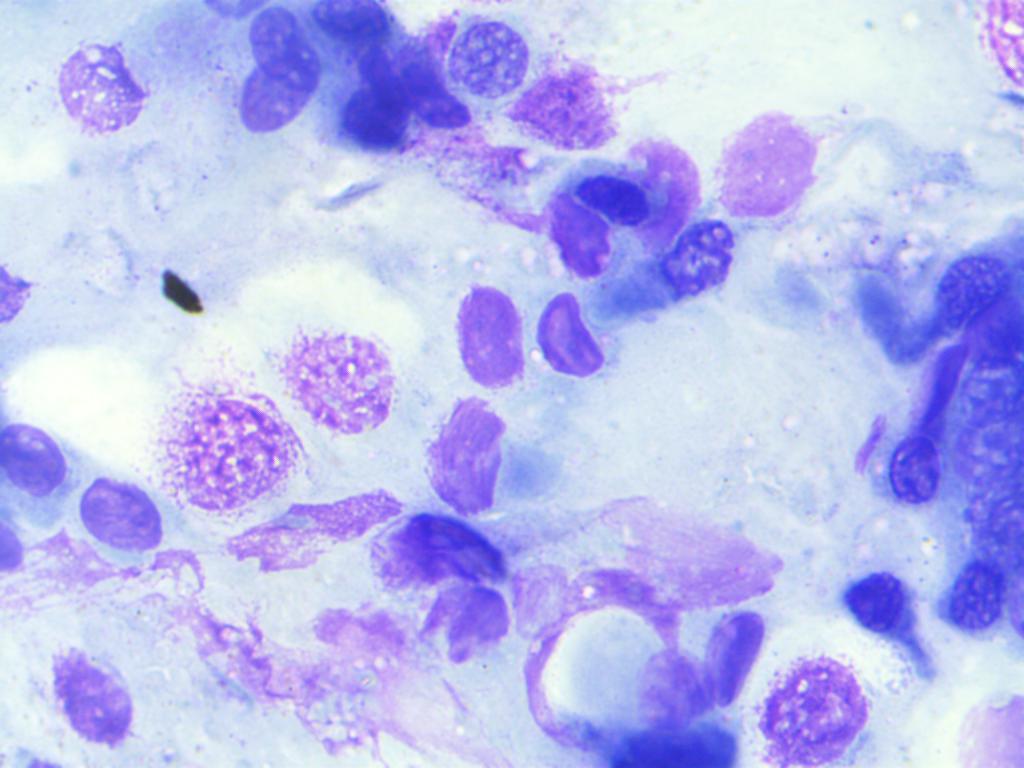}
         \caption{Cells covered with Biofilm }
         \label{2d}
     \end{subfigure}
    \caption{Examples of particularly OD difficult cases}
    \label{fig:three graphs} 
\end{figure}

Data were sampled from 14 rhinological slides collected at the Rhinology Clinic of the Otolaryngology Department of the University of Bari. Collecting technique was the direct smear and staining was the MGG. An optical microscope ProWay XSZPW208T  (figure \ref{2a}) with 1000x zoom, equipped with a  3MP DCE-PW300 camera (figure \ref{2b}) was used to acquire 50 images (microscope fields) from each slide: this specific quantity has been chosen since it is the one defined in the rhino cytology protocol. Thus 700 images with a size of 1024×768 were obtained. 

The image annotations were created by experts, using the Roboflow platform \cite{ROBOFLOW}, analyzing each image individually, annotating and labeling each cell. 

During such phase, a dropping policy was followed, discarding images whenever we detected: 
\begin{description}
    \item 1) sampling noise (i.e. dirt on the slide or blurred photos), 
    \item 2) duplication of large cytoplasmic areas already present in other images
    \item 3) Too dense and confused clusters of cells, typically discarded by nasal cytologist. 
\end{description}

A total of 200 cytological fields were pruned, ending up with 500 images.  
A Bounding Box (BB) was manually drawn on each cell in the images, to which a label was attached to specify the class the cell belonged to.  Being cells generally round, the smallest rectangular area that enclose them was marked as their bounding box. 

It is hence possible to find overlaps between BBs in images, owed by the proximity between the cells and the rectangular structure of the box. Labeling operations produced more than 10,000 BBs corresponding to cells. Thanks to Roboflow, annotations were made available in any standard annotation format required for computer vision algorithms, like Pascal Voc, Coco, Json and others.

 \begin{table}
    \centering
    \begin{tabular}{ |l||p{2cm}|p{2cm}|p{2cm}|p{2cm}|  }
        \hline
        \multicolumn{5}{|c|}{Number of occurrences} \\
        \hline
        Class &  TR  & VAL &TS & Total\\
        \hline
        \hline
         artifacts  & 820 & 98 & 100 & 1018\\
         emazia  & 37 & 6 & 5 & 48\\
          eosinophil & 420 & 54 & 54 & 528\\
        epithelial & 4062 & 503 & 495 &  5060\\
        epithelial ciliated & 93 & 11 & 11 &  115\\
        lymphocyte & 94 & 11 & 12 & 117\\
        mast-cell  & 15 & 2 & 2 & 19\\
        metaplastic  & 184 & 23 & 22 & 229\\
        muciparous & 403 & 51 & 50 & 504\\
        neutrophil & 2583 & 323& 308 & 3214\\

        \hline
    \end{tabular}
    \caption{Distribution of the 10 considered classes in the dataset.}
    \label{CELL-DIST-TABLE}
\end{table}

The 500 microscopic fields images were divided into training, validation and test set (80\%-10\%-10\%) using the stratified holdout strategy to maintain the same class distribution within the three sets. More details about class distribution can be found in Table \ref{CELL-DIST-TABLE}. 

\section{Methods} 

In order to provide a baseline as accurate as possible for the cell detection and cell recognition tasks, we chose to test and compare two different but equally high-performance architectures: the latest version of YOLO (YOLOv8) \cite{YOLOV8} and DETR \cite{DETR}, a transformer model specifically designed for object detection.

\subsection{YOLO v8}

The latest iteration of YOLO developed by Ultralytics \cite{YOLOV8} introduces new features and improvements to enhance the overall performance, flexibility and efficiency. This model supports a wide range of computer vision tasks, including object detection. The version of YOLOv8 used in this paper is YOLOv8x, the version with most parameters and thus more chances to learn to distinguish between the classes.

\subsection{The DETR model}

Developed \cite{DETR} with the base idea of processing the whole image and predict the entire set of bounding boxes and classification labels simultaneously (unique loss, computable in parallel for each image), avoiding explicit post process operations quite common in object detection, like duplicate suppression or anchoring, implementing an End-to-End model (direct set prediction). 

The architecture consists of a CNN backbone (in this study ResNet-50 \cite{RESNET50}), to embed visual features, an encoder and a decoder, both transformers based using multi head self-attention, which allow to always keep the whole image as context. The produced embeddings are fed to three ReLU-activated feed forward neural layers which respectively predicts center coordinates, height and width of the bounding boxes, along with their classes through one last softmax layer. The number of predicted bounding boxes $N$ is generally larger than the actual number of objects of interest, hence as many other OD models do, one additional "background" class is used to discard meaningless boxes. Training happens optimizing a set oriented cost function, called Hungarian Loss.

\subsection{Metrics}

To evaluate the performance of the two tested models, we rely on their specific evaluators, which report the results in term of Average Precision (AP) when referring to a single class and Mean Average Precision (mAP) when referring to multiple or all the cytotypes. In order to better grasp some concepts about AP, we introduce a few notes about Intersection over Union (IoU).

\begin{equation}
\label{eq_IoU}
    IoU = \frac{|A \cap B|}{|A \cup B|}
\end{equation}

\begin{figure}[htbp]
	\centering
    \begin{subfigure}[b]{0.30\textwidth}
    \centering
         \includegraphics[width=\textwidth]{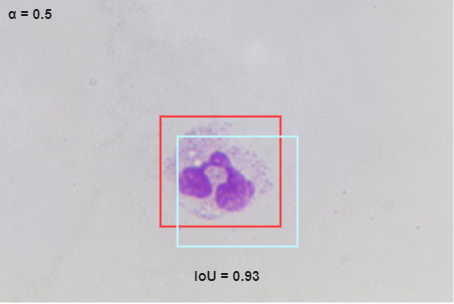}
         \caption{True Positive}
         \label{3a}
    \end{subfigure}
    \begin{subfigure}[b]{0.30\textwidth}
    \centering
         \includegraphics[width=\textwidth]{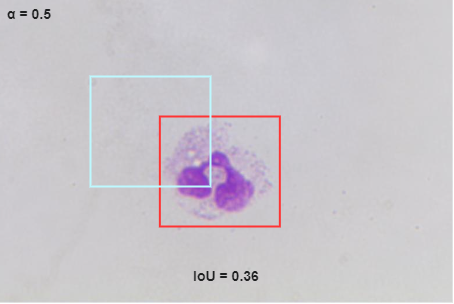}
         \caption{False Positive}
         \label{3b}
    \end{subfigure}
    \begin{subfigure}[b]{0.30\textwidth}
    \centering
         \includegraphics[width=\textwidth]{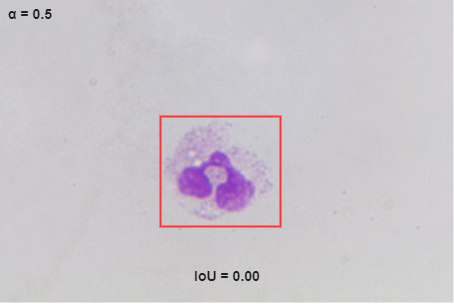}
         \caption{False Negative}
         \label{3c}
    \end{subfigure}
	\caption{Example of sorting for IoU threshold = 0.5}
	\label{fig:IoU}
\end{figure}

IoU (eq. \ref{eq_IoU}) quantifies the overlap between the prediction and the annotation by dividing the intersection area between the two bounding boxes ($A$ and $B$ in eq. \ref{eq_IoU}) by their union. An IoU threshold is defined to distinguish between True/False Positives/Negatives (figure \ref{fig:IoU}).

Evaluators for both models, compare the performances in terms of Average Precision (eq. \ref{eq_AP}).

\begin{equation}
\label{eq_AP}
    AP = \int_{0}^1 prec(rec) d(rec)
\end{equation}

Fixed an IoU threshold, precision and recall can be calculated and then a precision-recall curve is extracted. For both evaluators the AP is than calculated as the Area Under Curve (AUC) for the precision-recall curve.
We consider the AP50 which is the AUC of the precision-recall curve with 0.5 as IoU threshold and the AP50-95 which is the average between APs with IoU threshold varying from 0.5 to 0.95 with a step of 0.05.

\begin{equation}
\label{eq_mAP}
    mAP = \frac{1}{n} \sum_{k=1}^{k=n} AP_k
\end{equation}

Finally mean Average Precision (eq. \ref{eq_mAP}) is defined as the arithmetic average between the AP of each of the classes (in eq. \ref{eq_mAP} $n = 10$, as the number of classes).

\section{Experiments and Results}
\label{sec:experiments and results}

Two different experiments were carried out, both involving the two aforementioned models. 

The first experiment, referred as "Cell Recognition", aims at evaluating the performance of the model in the task of identifying and classifying cells into their respective cytotypes. As previously described, this task is very difficult even for specialists, given the strong inter-class similarity between some cytotypes and the high intra-class variance in the appearance of cells that belong to the same cytotype.

The second experiment focuses on establishing how well the two models can detect the cells in the cytologic field without assigning them to a specific cytotype. It will be referred from now on as "Cell Detection Task".

This additional task was thought in the optic of a generalized use of our dataset, as authors did for \cite{EVICAN} and \cite{LIVE-CELL}, developing OD models able to find cells independently on their staining, morphology and belonging tissue.

Observing pre-trained models finely learning to detect cells within nasal mucosa would hence make our dataset eligible to train some general purpose "Large Cell Detection Model" (LCDM).  

In particular, one of the "finest" skill the task requires to be accomplished is the  capability of dealing with dnese cells cluster \ref{2b}, very commonly occurring in nasal mucosa and are difficult to analyze even for expert rhinologists. From an OD point of view, such phenomena involves a large number of bounding boxes very densely distribute and usually overlapping. To provide a reasonable set of annotations both the overall and the local portion of the image should be taken into account, along with small details corresponding to the shape and the edges of each cell (membrane) \ref{2b}. Hence correctly learning or accomplishing the Cell Detection task on the NCD would confirm models to possess such OD skills.

Another point of view to motivate such a task is the possibility of splitting up the first task in two: (general) cell detection + cell classification, building up a specialized classifier on the top of a more versatile detector (similar to the approach used in \cite{NASAL-DL}).  

Each experiment was carried out in an experimental setup composed by an AMD Ryzen Threadripper PRO 3975WX with 32-Cores, a Nvidia Quadro RTX 5000 GPU with 16 GB of dedicated RAM and 64 GB of physical RAM.

\subsection{Cell Recognition}

The DETR model, pre-trained on the COCO dataset \cite{COCODATASET}, was fine-tuned for 50 epochs with our dataset splitted in train, validation and test set as described in section \ref{sec:dataset construction}. 



\begin{figure}[htbp]
	\centering
    \begin{subfigure}[b]{0.45\textwidth}
    \centering
         \includegraphics[width=\textwidth]{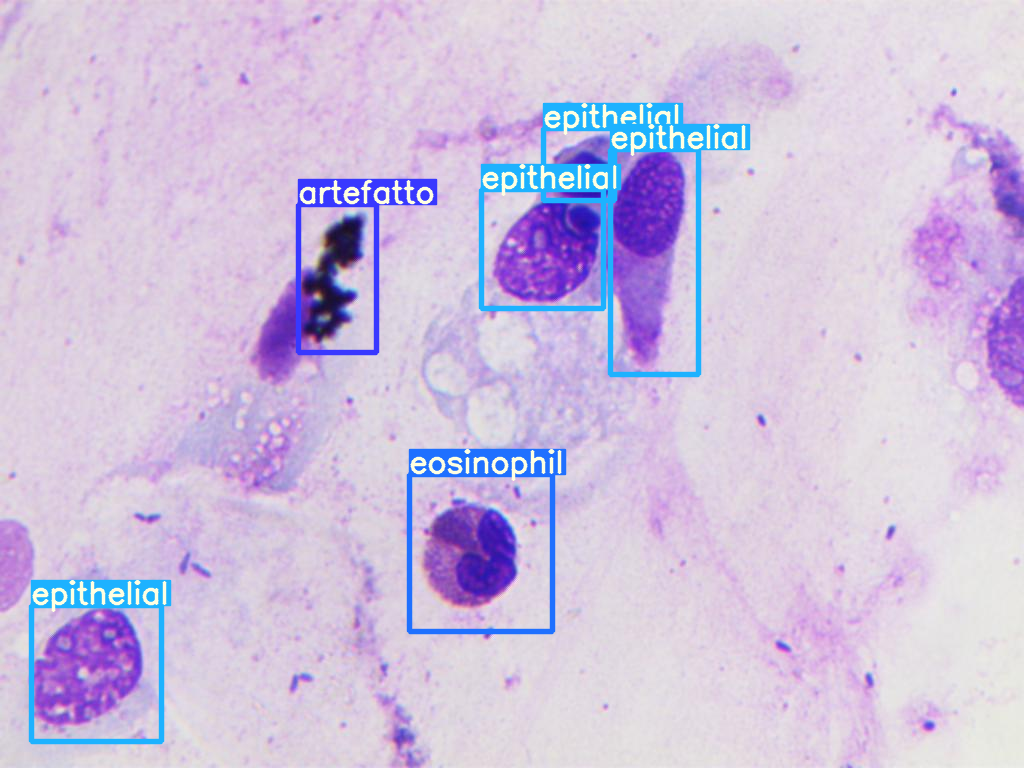}
         \caption{}
         \label{5a}
    \end{subfigure}
    \begin{subfigure}[b]{0.45\textwidth}
    \centering
         \includegraphics[width=\textwidth]{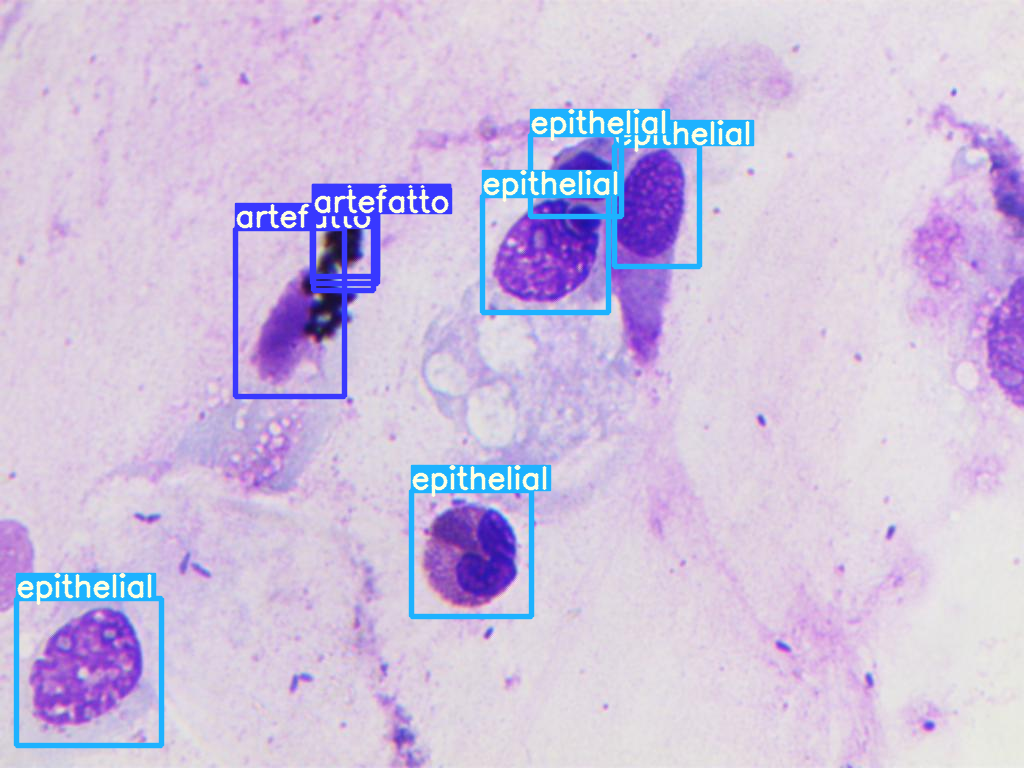}
         \caption{}
         \label{5b}
    \end{subfigure}
    \begin{subfigure}[b]{0.45\textwidth}
    \centering
         \includegraphics[width=\textwidth]{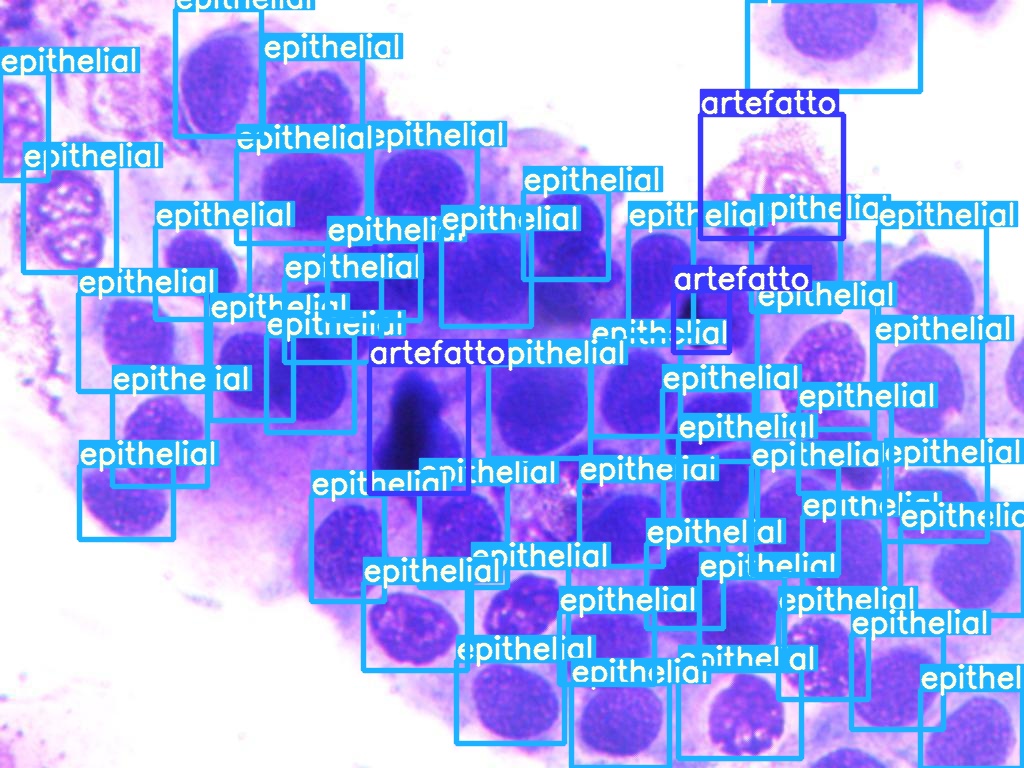}
         \caption{}
         \label{5c}
    \end{subfigure}
    \begin{subfigure}[b]{0.45\textwidth}
    \centering
         \includegraphics[width=\textwidth]{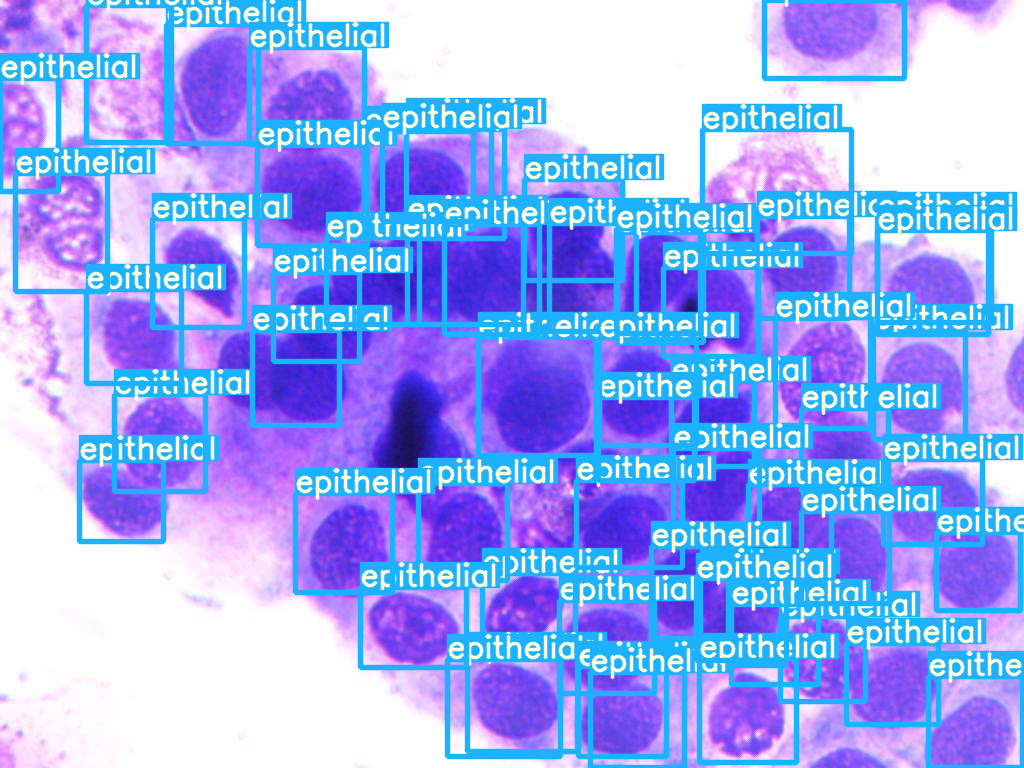}
         \caption{}
         \label{5d}
    \end{subfigure}
    \begin{subfigure}[b]{0.45\textwidth}
    \centering
         \includegraphics[width=\textwidth]{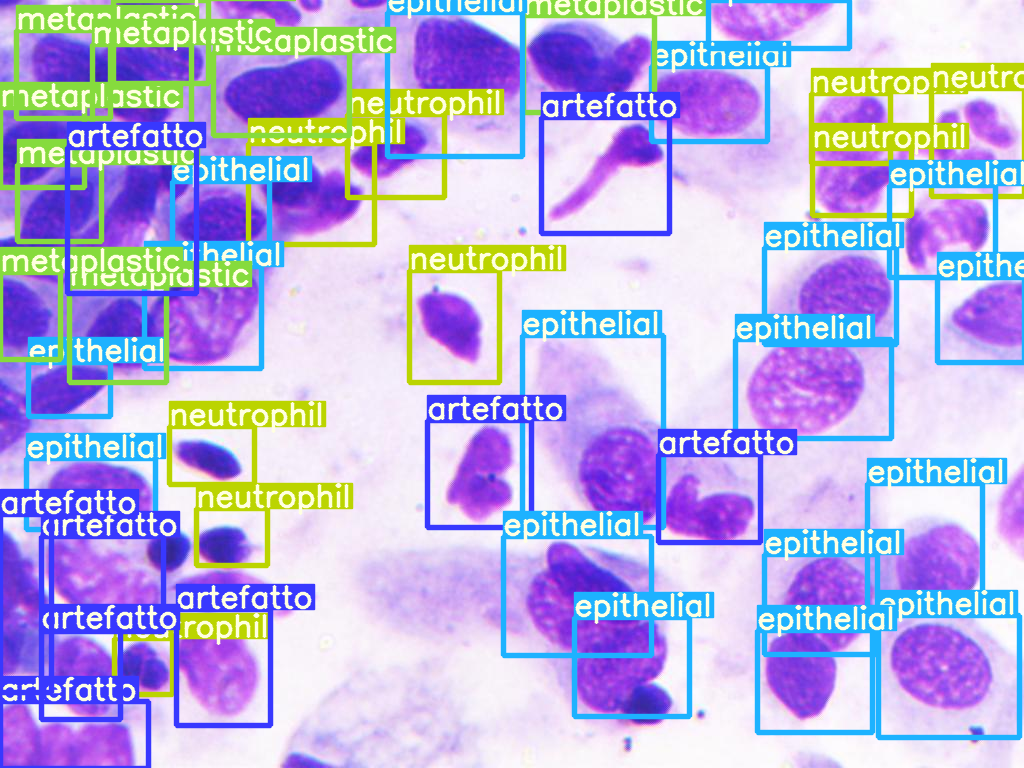}
         \caption{}
         \label{5e}
    \end{subfigure}
    \begin{subfigure}[b]{0.45\textwidth}
    \centering
         \includegraphics[width=\textwidth]{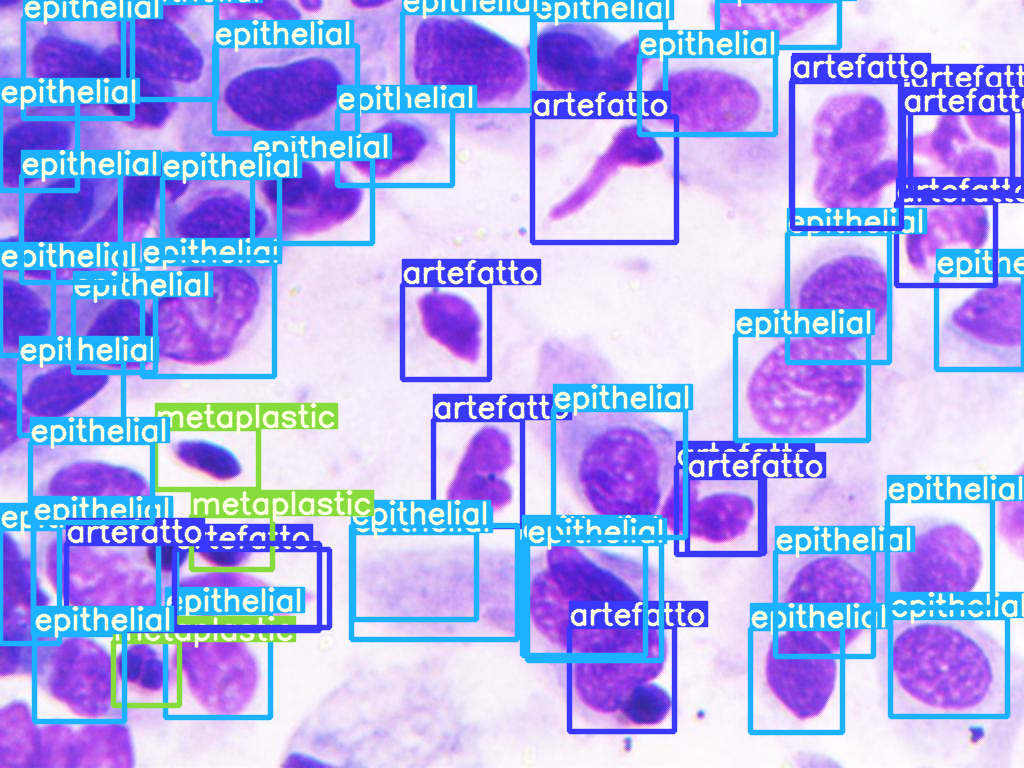}
         \caption{}
         \label{5f}
    \end{subfigure}
    \caption{Comparison between GTs (a,c,e) and DETR predictions (b,d,f) for the task of Cell Recognition.}
    \label{fig:DETR_identification_comparison}
\end{figure}

From figure \ref{fig:DETR_identification_comparison} one can assume that the model is clearly capable of detecting the cells (or artifacts) and assign them the correct label, but when comparing to the objective metrics presented in table \ref{table:detr_coco_identification} it can seem that there is no correspondence between qualitative results and quantitative ones.
These low results can be explained by two motivations.

The first is how restrictive is the Average Precision formula, that even if the classification of a cell is correct will not consider it if the IoU between the GT bounding box and the predicted one is lower than the selected IoU threshold; thus, is understandable how the difficulty of the human annotator in creating bounding boxes that are consistent can amplify the difference between GTs and predictions.

The second motivation comes from an intrinsic characteristic of the dataset presented in this study; since the samples distribution reflects the cytotypes in the nasal mucosa normal conditions, the dataset results unbalanced, a very frequent condition in medicine; this causes the model to become very good and confident in predicting classes where examples are numerous (section \ref{sec:dataset construction}) and very shy in predicting cytotypes with few samples. This in combination with the stratified splitting of the dataset results in some minority classes with less than 10 examples in the test set, that thus shows a very low AP. Being the mAP the mathematical average between the AP of all classes, these low scores greatly influence the final score.
An exception to this principle are eosinophils that can be easily distinguished from other cells and thus recognizable even if few examples are contained in the dataset.

\begin{table}[htbp]
\centering
\begin{tabular}{|l||p{2.5cm}|p{2.5cm}|p{2.5cm}|}
\hline
\textbf{Class}               & \textbf{Instances} & \textbf{mAP50} & \textbf{mAP50-95} \\ \hline \hline
artifact          & 100       & 0.133 & 0.062      \\ 
emazia              & 5         & 0     & 0    \\ 
eosinophil          & 54        & 0.402 & 0.180       \\ 
epithelial          & 495       & 0.738 & 0.337      \\ 
epithelial ciliated & 11        & 0.099 & 0.048       \\ 
lymphocyte          & 12        & 0     & 0     \\ 
mast cell           & 2         & 0     & 0          \\ 
metaplastic         & 22        & 0     & 0       \\ 
muciparous          & 50        & 0.055 & 0.013      \\ 
neutrophil          & 308       & 0.527 & 0.204      \\ \hline
\textbf{all}        & \textbf{1059} & \textbf{0.195} & \textbf{0.084}      \\ 
\hline
\end{tabular}
\caption{DETR Coco score on the test set for the task of Cell Recognition.}
\label{table:detr_coco_identification}
\end{table}

Subsequently, YOLOv8 model, which is also pre-trained on the COCO dataset, was fine-tuned for the same number of epochs with the same split of the previous experiment.

From table \ref{table: YOLO_results_Identification} the phenomenon described in the previous lines is clearly visible: the classes on which the model performs better (epithelial and neutrophil) are indeed the majority classes among all the cytotypes; we can thus observe how there is a visible correlation between mAP and representation of that classes in the dataset.
Exceptions are eosinophils that, as reported after the DETR experiment, are easily distinguishable from other cytotypes; this is also true for emazie that are blood stains in the cytologic slide. YOLO turns out to be better than DETR not only because it performs better in recognized cytotypes already identified by the former model, but has also because it learned to recognize 3 more classes that were not distinguished by DETR.

The solution to this unbalancing problem is not trivial; it is true that applying a simple weighted average instead of the arithmetic one will result in higher overall accuracy, but that will not make better in any way the qualitative results of the system, but will instead give us the impression that the system is performing very well, while is indeed only ignoring the minority classes, among which there are some that are equally, if not more, important at diagnostic level to the more numerous ones.

\begin{table}[htbp]
\centering
\begin{tabular}{|l||p{2.5cm}|p{2.5cm}|p{2.5cm}|}
\hline
\textbf{Class}               & \textbf{Instances} & \textbf{mAP50} & \textbf{mAP50-95} \\ \hline \hline
artifact           & 100        & 0.293 & 0.171      \\ 
emazia              & 5         & 0.831 & 0.426    \\ 
eosinophil          & 54        & 0.853 & 0.528       \\ 
epithelial          & 495       & 0.863 & 0.488      \\ 
epithelial ciliated & 11        & 0.306 & 0.196       \\ 
lymphocyte          & 12        & 0.223 & 0.125     \\ 
mast cell           & 2         & 0     & 0          \\ 
metaplastic         & 22        & 0.051 & 0.032       \\ 
muciparous          & 50        & 0.517 & 0.251      \\ 
neutrophil          & 308       & 0.843 & 0.440      \\ \hline
\textbf{all}        & \textbf{1059} & \textbf{0.478} & \textbf{0.266}      \\ 
\hline
\end{tabular}
\caption{YOLOv8 scores on the test set for the task of Cell Recognition.}
\label{table: YOLO_results_Identification}
\end{table}

\begin{figure}[htbp]
	\centering
    \begin{subfigure}[b]{0.45\textwidth}
    \centering
         \includegraphics[width=\textwidth]{IMAGES/6/2o.png}
         \caption{}
         \label{6a}
    \end{subfigure}
    \begin{subfigure}[b]{0.45\textwidth}
    \centering
         \includegraphics[width=\textwidth]{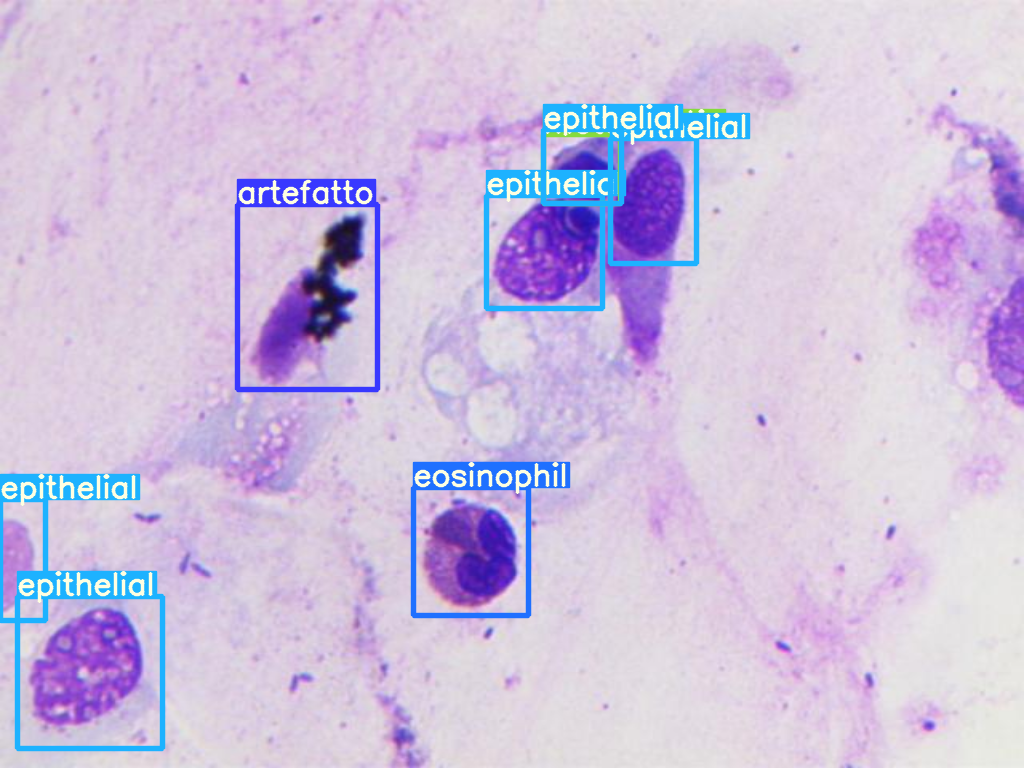}
         \caption{}
         \label{6b}
    \end{subfigure}
    \begin{subfigure}[b]{0.45\textwidth}
    \centering
         \includegraphics[width=\textwidth]{IMAGES/6/1o.png}
         \caption{}
         \label{6c}
    \end{subfigure}
    \begin{subfigure}[b]{0.45\textwidth}
    \centering
         \includegraphics[width=\textwidth]{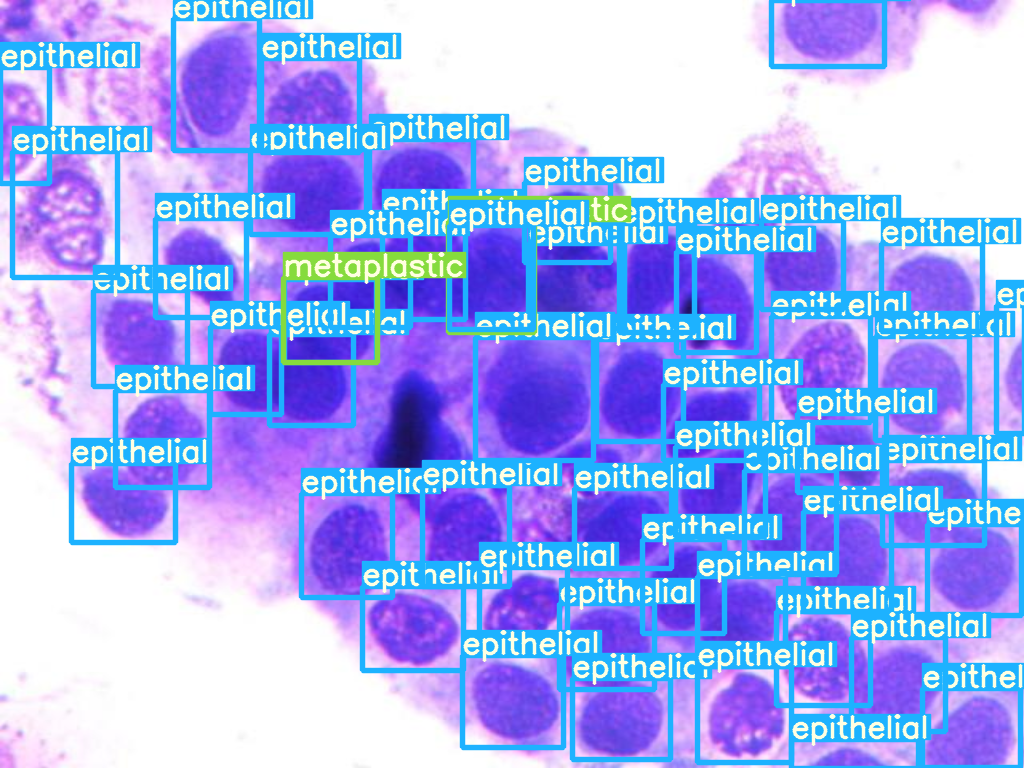}
         \caption{}
         \label{6d}
    \end{subfigure}
    \begin{subfigure}[b]{0.45\textwidth}
    \centering
         \includegraphics[width=\textwidth]{IMAGES/6/3o.png}
         \caption{}
         \label{6e}
    \end{subfigure}
    \begin{subfigure}[b]{0.45\textwidth}
    \centering
         \includegraphics[width=\textwidth]{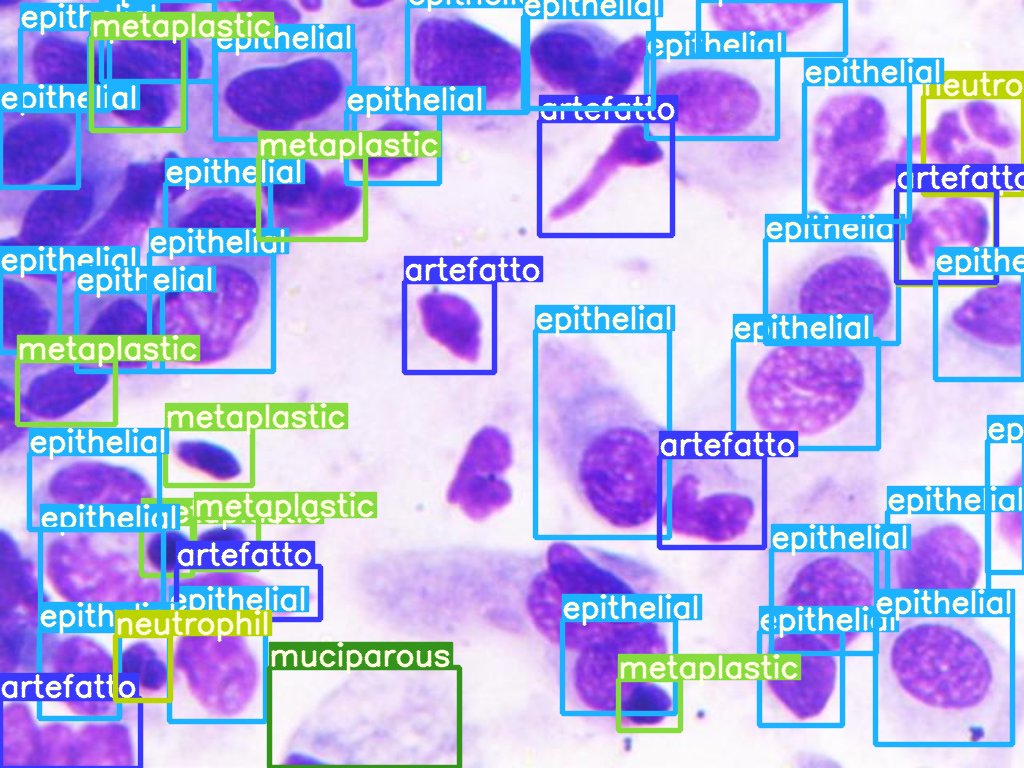}
         \caption{}
         \label{6f}
    \end{subfigure}
    \caption{Comparison between GTs (a,c,e) and YOLOv8 predictions (b,d,f) for the task of Cell Recognition.}
    \label{fig:YOLO_identification_comparison}
\end{figure}

\subsection{Cell Detection}
To investigate how well the two models can focus on detecting the presence of cells and how precise are the predicted BBs (Cell detection task), another experiment was carried out.

To do so, a duplicated version of the dataset was generated. Being not actual cells, all the labels concerning the classes "artifact" and "emazia" were removed and thus considered background. Then, all the remaining labels were renamed and grouped into a single class named "cell", maintaining their correspondence with the original bounding boxes.

As previously mentioned, aim of such experiment is very whether models are able to deal with general "cells" objects correctly detect them even within complex dense clusters, independently on their type.

As in the previous experiment, the first model to be tested was DETR that, also in this case, was fine-tuned by 50 epochs, followed by the YOLO fine-tuning that was also identical to the previous experiment.

\begin{table}[htbp]
\centering
\begin{tabular}{|l||p{2.5cm}|p{2.5cm}|p{2.5cm}|p{2.5cm}|}
\hline
\textbf{Model} & \textbf{Class} & \textbf{Instances} & \textbf{AP50} & \textbf{AP50-95} \\ \hline \hline
DETR & Cell & 978 & 0.780 & 0.339  \\
YOLOv8 & Cell & 978 & 0.888 & 0.505      \\ 
\hline
\end{tabular}
\caption{DETR and YOLOv8 scores on the test set for the task of Cell Detection.}
\label{table: results_detection}
\end{table}

\begin{figure}[htbp]
	\centering
    \begin{subfigure}[b]{0.45\textwidth}
    \centering
         \includegraphics[width=\textwidth]{IMAGES/8/1o.png}
         \caption{}
         \label{8a}
    \end{subfigure}
    \begin{subfigure}[b]{0.45\textwidth}
    \centering
         \includegraphics[width=\textwidth]{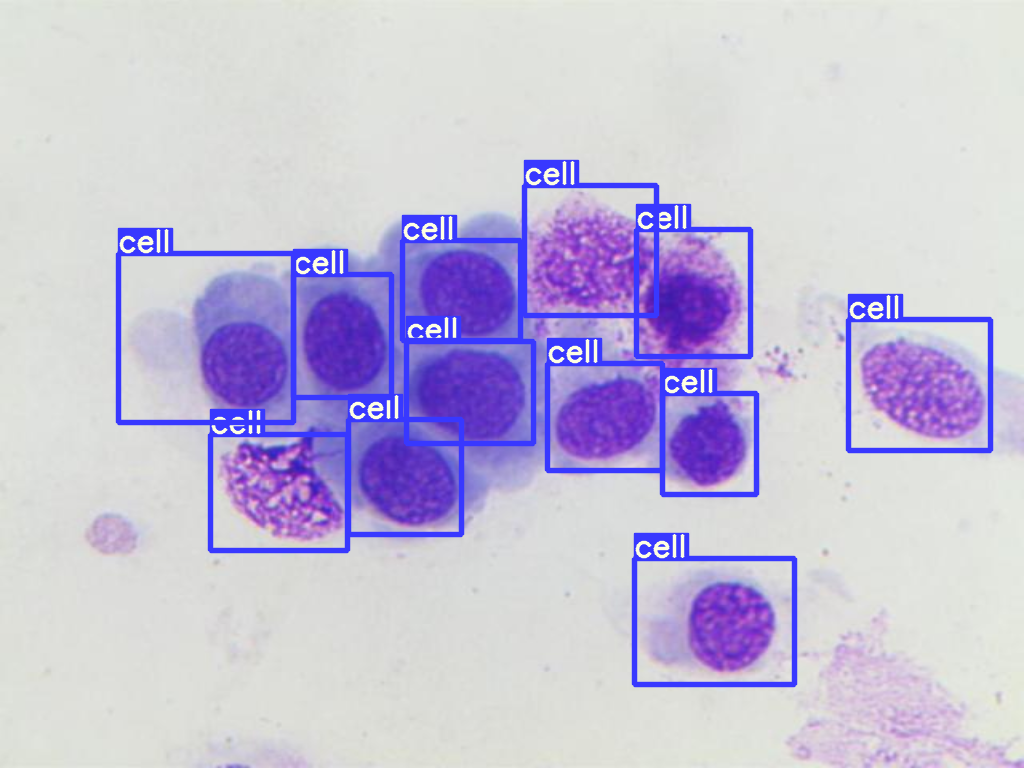}
         \caption{}
         \label{8b}
    \end{subfigure}
    \caption{Comparison between a GT (a) and the respective YOLOv8 prediction (b) for the task of Cell Detection.}
    \label{fig:YOLO_detection_comparison}
\end{figure}

More optimistic, yet predictable results can be inferred from tables \ref{table: results_detection}: the task of cell identification is very well addressed by the two models, especially YOLO. Since we are not interested in perfect correspondence between the predicted bounding box and the manually annotated one, we can be satisfied by the AP50 obtained by YOLO that reaches 87\%, a result that is even more encouraging when looking at figure \ref{fig:YOLO_detection_comparison} where annotations are sometimes even more centered and focused on the cell than the ground truths.

Although satisfying, the results are still improvable: even though both models are capable to identify the great majority of cells, they tend to be too much inclusive with the learned definition of cell. This results in some cases where "artifacts" or even "emazie" are classified as cells.

\begin{figure}[htbp]
	\centering
    \begin{subfigure}[b]{0.45\textwidth}
    \centering
         \includegraphics[width=\textwidth]{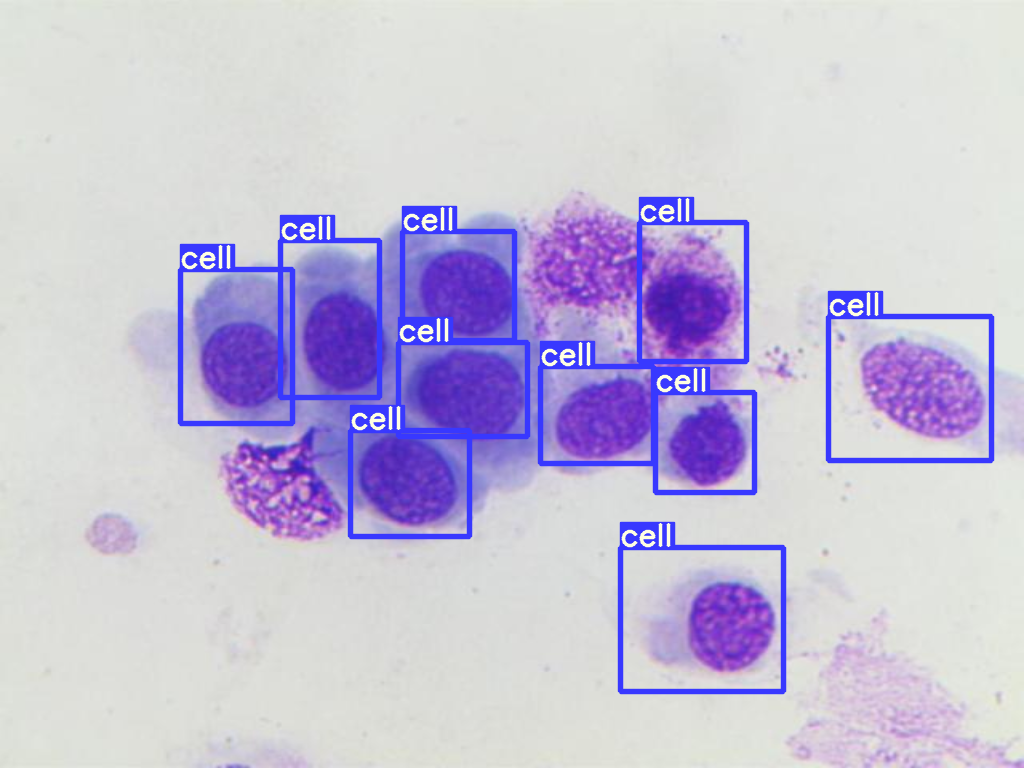}
         \caption{}
         \label{7a}
    \end{subfigure}
    \begin{subfigure}[b]{0.45\textwidth}
    \centering
         \includegraphics[width=\textwidth]{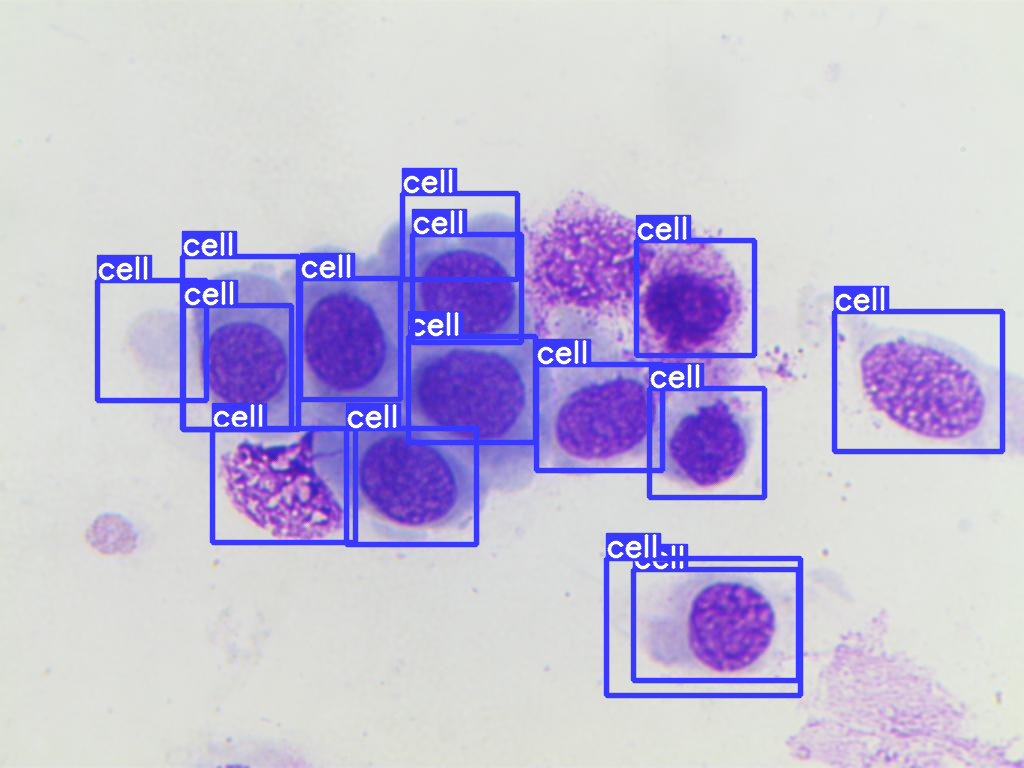}
         \caption{}
         \label{7b}
    \end{subfigure}
    \caption{Comparison between a GT (a) and the respective DETR prediction (b) for the task of Cell Detection.}
    \label{fig:DETR_detection_comparison}
\end{figure}

\section{Discussion}
\label{sec:discussion}

Some aspects of the results showed in the previous section are worth to be better discussed and analyzed, in the prospective of defining which are the challenges that are characteristic to the task of automatic nasal mucosa cell recognition.

The first aspect to be discussed is how, the two models evaluated are surely capable of detecting cells in the pictures of cytological field, capturing them in their entirety; this was expected since other cytological studies have already applied this two (or similar) models in task that are not too much dissimilar from the ones analyzed in this study \cite{TRANSFER-LEARNING-CERVIX-CYTO, DL-CYTOLOGY-SURVEY, LIQUID-YOLO}, obtaining good result.

Unfortunately this is not enough to help clinicians, since the total number of cells has no diagnostic information; what we are interested in is the number of cells for each of the 8 cytotypes ("emazia" and "artifact" are not cells) but, as observed in the experimental results, the two models are not capable yet of giving us correct information about the number of cells for each cytotype; this mainly happens because the proposed dataset has a realistic cell distribution that makes some classes greatly over represented while other nearly not represented at all.

This leads to results, for the cell classification task, which
are themselves unbalanced: accurate for the over/well-represented classes and bordering on zero for the minority classes.

The diagnostic significance of a cytotype is independent of its standard quantity within the nasal mucosa, but is more dependent to what is the cell specific task, with significance proportional to the percentage of variation between the actual distribution and the standard one; thus it becomes crucial to investigate how to balance the dataset to bring the models to a less biased and more objective classification.

Methods that could be investigated to solve this type of problem could be the use of a stable diffusion model that allows the dataset to be augmented with artificial examples containing only or mainly minority classes, but this method would require an expert to confirm that the artificially generated images are medically credible.

Another viable approach would be to divide the two steps (detection and classification), so that augmentation can be performed on individual cells, thus increasing the number of examples of a cytotype without affecting the others.

Another aspect that is worth discussing is how the models manage cell clusters; when physiologist are manually counting cells, to avoid to lose a lot of time identifying and counting each cell in a cluster, they decide to consider that all the cells in the cluster belong to a specific cytotype (the one that the majority of cells in that cluster belong to) and add an arbitrary but plausible number of cells to the count for that cytotype.
When annotations for the dataset were created, it was specifically chosen to identify single cells even in clusters; it was a very hard and time consuming work, but it would allow to train models to learn to recognize cells even when clustered.

This can be indeed seen in figures \ref{5d} and \ref{6d} where it is shown how both models are capable of identifying a good number of clustered cells.
Counting cells in a cluster and more importantly distinguish their cytotype instead of assuming that are all from the same one, can surely guarantee more precise diagnostic results.

\section{Conclusions}
\label{sec:conclusion}

Nasal cytology has proven to be an efficient and inexpensive technique for the diagnosis of rhinitis and allergies, however, it is not widely used because of how long the cell count procedure takes. With this work, we therefore intend to initiate and stimulate a new research AI research field: that on the automatic detection of nasal mucous membrane cells, so that, with a rapid and autonomous tool, the spread of this diagnostic practice can accelerate.
To do this, in this paper we present a new dataset that is the result of years of work, collecting and labelling photographs from rhinocytologic fields: slides of nasal mucosa smears. This is the first dataset in its field and will make it possible to allow researchers to experiment new OD, segmentation and classification models, or even to try transferring knowledge from other tissues or domains. Such last aspect is particularly promising to evaluate model invariance with respect to staining, microscope zoom and lightening checking their robustness for clinical practice \cite{LIQUID-YOLO}. From the point of view of AI, we fine-tuned two Deep Learning models with different architectures, DETR and YOLOv8. Both have proven to be able to detect cells without difficulty, even in cell clusters, and to classify them accurately when dealing with common cytotypes, and with some uncertainty when analysing rarer cytotypes; the results obtained from this experiment are intended to serve as a baseline for the rest of the scientific community to have values against which to compare for possible other solutions for this complex and unbalanced task, which would have been critical for the majority of existing OD models. 
 

\printbibliography 
\end{document}